%% file: main.tex
\documentclass[sigconf]{acmart}

\usepackage{booktabs} 
\usepackage{balance} 

\begin{document}
\title{PatternNet: Visual Pattern Mining with Deep Neural Network}

\author{Hongzhi Li \footnotemark[1],  Joseph G. Ellis \footnotemark[2], Lei Zhang \footnotemark[1] and Shih-Fu Chang \footnotemark[2]}
\affiliation{%
  \institution{\footnotemark[1] Microsoft Research,  Redmond, WA 98052, USA}
}
\affiliation{%
  \institution{\footnotemark[2] Columbia University, New York, NY 10027, USA}
}
\email{hongzhi.li@microsoft.com, jge2105@columbia.edu, leizhang@microsoft.com, sfchang@ee.columbia.edu}

\begin{abstract}
Visual patterns represent the discernible regularity in the visual world.
They capture the essential nature of visual objects or scenes.
Understanding and modeling visual patterns is a fundamental problem in visual recognition that has wide ranging applications.
In this paper, we study the problem of visual pattern mining and propose a novel deep neural network architecture called PatternNet for discovering these patterns that are both discriminative and representative.
The proposed PatternNet leverages the filters in the last convolution layer of a convolutional neural network to find locally consistent visual patches, and by combining these filters we can effectively discover unique visual patterns.
In addition, PatternNet can discover visual patterns efficiently without performing expensive image patch sampling, and this advantage provides an order of magnitude speedup compared to most other approaches.
We evaluate the proposed PatternNet subjectively by showing randomly selected visual patterns which are discovered by our method and quantitatively by performing image classification with the identified visual patterns and comparing our performance with the current state-of-the-art.
We also directly evaluate the quality of the discovered visual patterns by leveraging the identified patterns as proposed objects in an image and compare with other relevant methods.
Our proposed network and procedure, PatterNet, is able to outperform competing methods for the tasks described.
\end{abstract}

%
%

\copyrightyear{2018} 
\acmYear{2018} 
\setcopyright{acmlicensed}
\acmConference[ICMR '18]{2018 International Conference on Multimedia Retrieval}{June 11--14, 2018}{Yokohama, Japan}
\acmBooktitle{ICMR '18: 2018 International Conference on Multimedia Retrieval, June 11--14, 2018, Yokohama, Japan}
\acmPrice{15.00}
\acmDOI{10.1145/3206025.3206039}
\acmISBN{978-1-4503-5046-4/18/06}


\keywords{visual pattern mining; convolutional neural network; image classification; object proposal}

\maketitle
\input{body}


\bibliographystyle{abbrv}
\balance 
\bibliography{main}

\end{document}

%% file: body.tex
\section{Introduction}

\begin{figure*}
  \centering
  \includegraphics[width=0.9\textwidth]{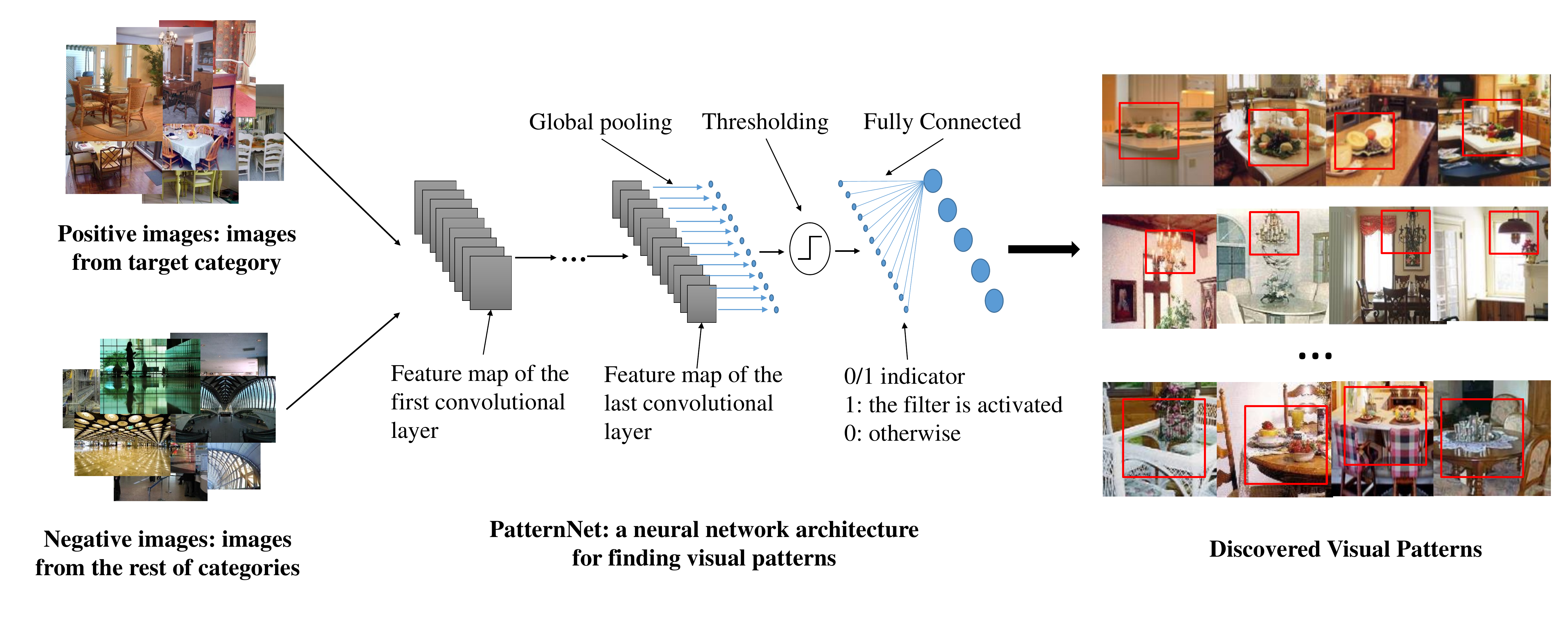}
  \caption{Visual Pattern Mining with Deep Neural Network}
  \label{fig:patternmining}
\end{figure*}


Visual patterns are basic visual elements that commonly appear in images but tend to convey higher level semantics than raw pixels. Visual patterns are a reflection of our physical world, in which plants and animals reproduce themselves by repeated replication and we as human being improve our world by constant innovation and duplicative production. As a result, we see in images many similar and repeated patterns at different semantic levels, for example, lines, dots, squares, wheels, doors, cars, horses, chairs, trees, etc.

Understanding and modeling visual patterns is a fundamental problem in visual recognition, and much work has been done in the computer vision community to address this problem, with varied approaches and success. For example, SIFT features can be used to detect robust local points that are scale-invariant and can tolerate limited distortions \cite{lowe1999object}. The detected SIFT patches are often regarded as low level image patterns.

The convolutional layers in a Convolutional Neural Network can be seen as a form of feature extractor or visual pattern mining. 
CNNs have recently been shown to exhibit extraordinary power for visual recognition. The breakthrough performance on large-scale image classification challenges \cite{krizhevsky2012imagenet,Simonyan14c} is just one example.
Many researchers recently have been interested in understanding why CNNs exhibit such strong performance in classification tasks and have been analyzing the changes in the structure and underlying theory of CNNs. One of the the most popular interpretations \cite{zeiler2014visualizing}  is that the trained convolution layers are able to capture local texture patterns of the image set.
\cite{zeiler2014visualizing} designed a deconvolutional neural network to visually demonstrate the information captured by each convolutional filter in a convolutional neural network \cite{krizhevsky2012imagenet}. Given any filter, the deconvolution process traces back via the network and finds which \textit{pixels} in the original image contribute to the response of this filter. Using the deconvolution neural network, one can show that each filter is normally sensitive to certain visual appearances or patterns and can demonstrate what type of patterns each filter is sensitive to. 
For example, the first one or two convolution layers are able to capture simple textures like lines or corners, whereas the upper layers are capable of capturing semantically meaningful patterns with large variances in appearance. 
A typical CNN, like AlexNet, has 256 filters in its last convolutional layer (conv5), which is a very small number compared with all the possible patterns existing in the real world.
This implies that a filter may be triggered by several different patterns that share the same latent structure that is consistent with the filter.
We also find that an image patch can trigger several different filters simultaneously when it exhibits multiple latent patterns that conform with multiple filters.

Base on above analysis of CNN filters, we study the problem of visual pattern mining and propose a framework for leveraging the activations of filters in a convolutional neural network to automatically discover patterns.
However, filters and activations from CNN architectures as currently constructed can not be used directly to find visual patterns. Therefore, we propose a new network structure designed specifically to discover meaningful visual patterns.

Filters can be triggered by image patches from the same visual pattern, but for our definition of a visual pattern a set of similar image patches must be 1) popular and 2) unique.
Formally, each of our discovered patterns should be \textbf{representative} and \textbf{discriminative}. Discriminative means the patterns found from within one image category should be significantly different from those that are found in other categories, and therefore only appear sparingly in those other categories.
This means that patterns should represent unique visual elements that appear across images in the same category. Representative requires that the patterns should appear frequently among images in the same category.
That is, a visual pattern should not be an odd patch only appearing in one or two images.
Patterns that do not appear frequently may be discriminative but will not appear in enough images to be of use.

We formulate the problem  of visual pattern mining as follows: given a set of images from a category (images from the target category are referred to as ``positive images'' throughout this manuscript), and a set of images from other categories as reference (these are referred to as ``negative images'' in the rest of this paper), find representative and unique visual patterns that can distinguish positive images from negative images.
Our discriminative property insures that the patterns we find are useful for this task.
If a pattern only appears in positive images but not in negative images, we call it discriminative.
If a pattern appears many times in positive images, we call it representative.

In this paper, We follow our two defined criteria for discovering visual patterns (representative and discriminative) and design a neural network inspired by the standard convolutional neural network used for image classification tasks.
We name the proposed neural network \textbf{PatternNet}. PatternNet leverages the capability of the convolution layers in CNN, where each filter has a consistent response to certain high level visual patterns.
This property is used to discover the discriminative and representative visual patterns using a specially designed fully connected layer and loss function to find a combination of filters which have strong response to the patterns in the images from the target category and weak response to the images from other categories.After we introduce the architecture of PatternNet, we will analyze and demonstrate how PatternNet is capable of finding the representative and discriminative visual patterns from images.

The contributions of this paper are highlighted as follows:
\begin{itemize}
\item We propose a novel end-to-end neural network architecture called PatternNet for finding high quality visual patterns that are both discriminative and representative.
\item By introducing a global pooling layer between the last convolutional layer and the fully connected layer, PatternNet achieves the shift-invariant property on finding visual patterns. This allows us to find visual patterns on image patch level without pre-sampling images into patches.
\end{itemize}

\section{Related Work}
Visual pattern mining and instance mining are fundamental problems in computer vision.
Many useful and important image understanding research and applications rely on high-quality visual pattern or visual instance mining results, such as the widely used mid-level feature representations for image classification and visual summarization.
Most previous works \cite{singh2012unsupervised,juneja2013blocks,li2013harvesting,sun2013learning} follow the same general procedure. First, image patches are sampled either randomly from images or by using object proposals, such as selective search, saliency detection, or visual attention. Then visual similarity and geometry restrictions are employed for finding and clustering visually similar image patches, which are often referred to as \textit{visual patterns}.
Subsequently, the discovered visual patterns are used to build a mid-level feature representation that can improve the performance of image classification.

\begin{figure*}
  \centering
  \includegraphics[width=0.9\textwidth]{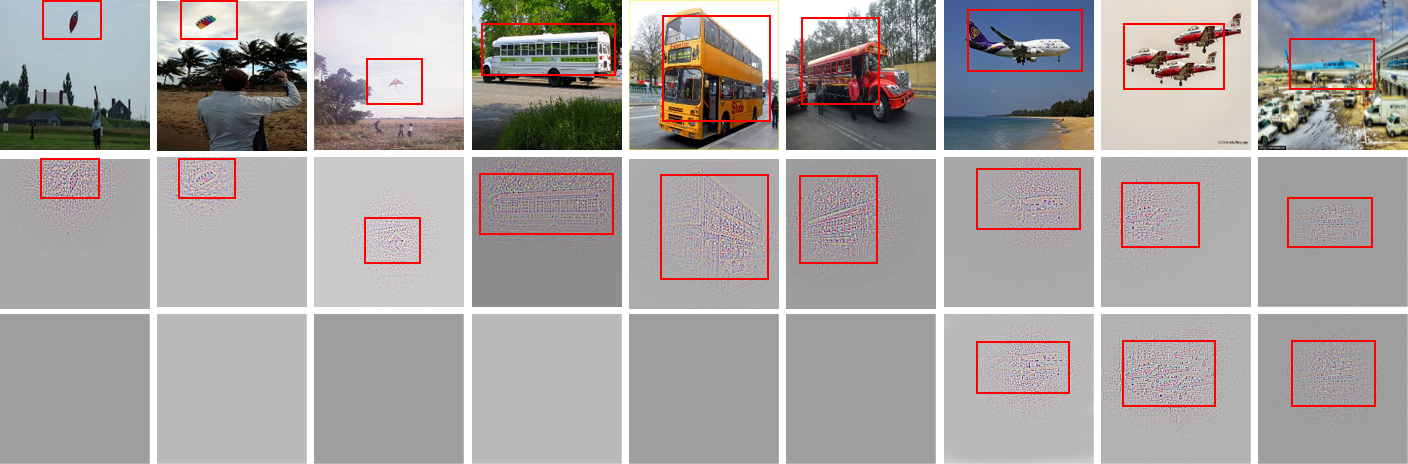}
  \caption{\small{Visualization of the local response region of filters in the last convolutional layer of CNN on different images. The second row images show the local response region of filter $\alpha$. Filter $\alpha$ is activated by all the images. The third row images show the local response region of filter $\beta$. Filter $\beta$ is only activated by the ``flight'' images}}
  \label{fig:deconv_filters}
\end{figure*}

As well as using visual patterns as middle-level feature representations, visual pattern mining itself can be used in broader research areas. Zhang et al.  \cite{zhang2014scalable} propose a method to use ``thread-of-feature'' and ``min-Hash'' technology to mine visual instances from large-scale image datasets and apply the proposed method on the applications of multimedia summarization and visual instance search.

Some works, such as \cite{parizi2014automatic} and \cite{krause2015fine} use the term ``parts'' to describe a similar concept to ``visual patterns'' in this paper. They define ``part'' as a partial object or scene that makes up a larger whole. In part-based approaches, the objects or scenes are broken into parts and a binary classifier is trained for each part. The parts are used as an image representation for image classification. Parts-based image classification works focus on different aspects than our work. First, those works are supervised approaches. The part detectors are learned from labeled data, while PatternNet uses unsupervised learning techniques to find the visual patterns from a set of images. Second, the goal of using parts-based models is to obtain better classification results, while we focus on finding discriminative and representative visual patterns.

More recently, \cite{lilsh15cvpr} has utilized a Convolution Neural Network for feature representation and has used association rule mining \cite{agrawal1993mining}, a widely used technique in data mining,  to discover visual patterns.
The key idea of this approach is to form a transaction for each image based on its neuron responses in a fully connected layer and find all significant association rules between items in the database.

In contrast to most existing approaches that normally have a separate stage to extract patches or construct transactions, followed with a clustering algorithm or an association rule mining algorithm for finding useful visual patterns, we develop a deep neural network framework to discover visual patterns in an end-to-end manner, which enables us to optimize the neural network parameters more effectively for finding the most representative and discriminative patterns.

\section{Visual Pattern Mining}
In this section, we introduce the architecture and analyze the properties of our novel CNN, PatternNet.

\begin{figure*}
  \centering
  \includegraphics[width=0.75\textwidth]{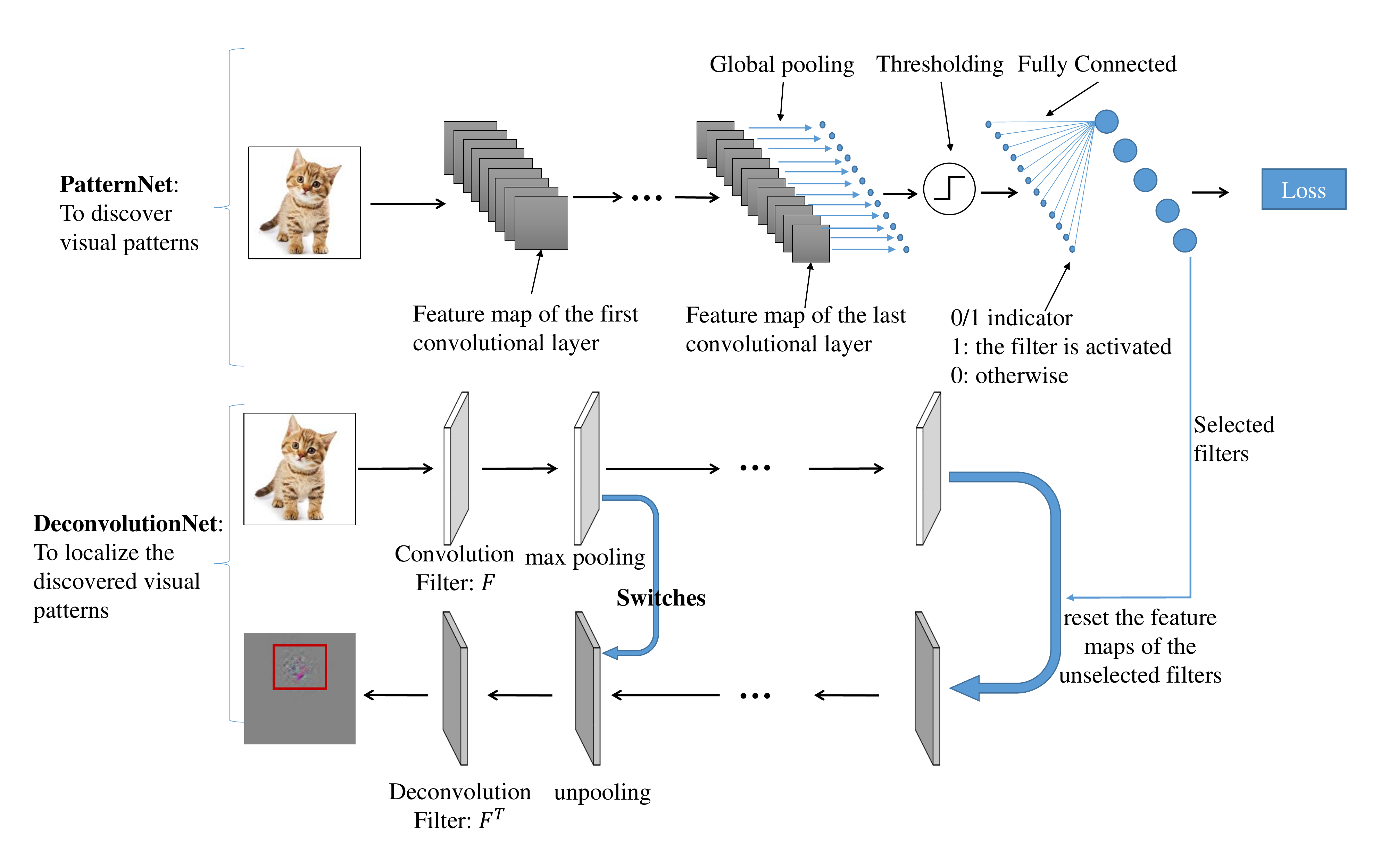}
  \caption{\small{PatternNet Architecture. The proposed system use PatternNet to discover visual patterns. The visual patterns are represented by a set of filters. Once the visual patterns are discovered, a deconvolutional neural network is used to localize the location of the visual pattern in the input image.}}
  \label{fig:patternnet}
\end{figure*}

\subsection{PatternNet Architecture \label{sec:patternnet}}
The convolutional layers in CNNs can be seen as feature extractors. An excellent visual example demonstrating the capabilities of each CNN as a feature extractor exists in \cite{zeiler2014visualizing}.
They demonstrate that the first few convolutional layers tend to capture lower-level image features, such as edges and corners, while the last convolutional layer captures higher-level object information, such as people's faces, wheels, or other complicated structural patterns. Recently, the properties of convolutional layers have been leveraged to address the problem of object segmentation and have shown very promising results in \cite{hariharan2015hypercolumns}.
The activations of convolutional layers can be applied in object segmentation problems because: 1) a filter is normally selective to certain patterns, and 2) a filter is spatially local and its response map can be utilized to localize the image patch that has the object-of-interest in the original input image.
We will leverage both of these tendencies, especially the ability for a convolutional filter to provide local information within the image in our construction of PatternNet.

As shown in Figure \ref{fig:deconv_filters}, we visualize the local response region of the same filter in the last convolutional layer of a CNN on different images. We can clearly see that the filter can be activated by different visual patterns. Due to the fact that a filter may be activated by multiple different visual patterns, we cannot directly use the single filter output to discover these visual patterns. On the other hand, a visual pattern may activate multiple filters, as well.

Thusly, we can think of the filters in the last convolutional layer as more like selectors for finding mid-level \textit{latent} visual patterns rather than high-level \textit{semantic} visual patterns.

This observation motivates us to develop a neural network architecture to find visual patterns as a combination of multiple filters from the final convolutional layer. For example, in Figure \ref{fig:deconv_filters}, the visual pattern ``flight'' can be detected by the filters $\{\alpha, \beta \}$.
A typical convolutional neural network has $N_c$ filters in its last convolutional layer, and each filter produces an $M_c \times M_c$ dimensional feature response map.
The value of each pixel in a feature map is the response of a filter with respect to a sub-region of an input image.
A high value for a pixel means that the filter is activated by the content of the sub-region of the input image.
If a filter is activated, the $M_c \times M_c$ dimensional feature map records the magnitude of the response and the location information of where in the input image the filter is activated.
We use a tunable threshold $T_r$ to decide whether the response magnitude is sufficiently high such that the filter should be considered activated.
We set the response as 1 if the response magnitude is stronger than $T_r$, or $0$ otherwise.
To achieve translation-invariance and more effectively utilize image sub-regions, we intentionally ignore the location information. That is, as long as there is at least one response value from the $M_c \times M_c$ feature map larger than $T_r$, we consider that the filter is activated by the input image.

In PatternNet, we use a global max pooling layer after the last convolutional layer to discard the location information, which leads to a faster and more robust pattern mining algorithm as the feature after global max pooling is more compact and can effectively summarize input patches from any location. We utilize the deconvolutional neural network to obtain the location information later in the process when we need to localize the visual patterns.
Each feature map produces one max value after the global pooling layer. This value is then sent to a threshold layer to detect whether the corresponding filter is activated by the input image.
After thresholding, we get $N_c$ of 0/1 indicators for $N_c$ filters in the last convolution layer. Each indicator represents if a filter is activated by the input image.
We use a fully connected (FC) layer to represent the selection of filters for the visual patterns.
Each neuron in the fully connected layer is a linear combination of the outputs from the global pooling layer after thresholding:
\small{
\begin{equation}
h_{i} = \sum_{k=1}^{N_c}{W_{i,k} \cdot x_{k}}
\end{equation}
}
where $h_{i}$ is the response of the $i$-th $(i=1,...,N_f)$ neuron in the FC layer, and $x_{k} \in \{0,1\}$ is the activation status of the $k$-th filter in the last convolutional layer.
The selection of filters is reflected by the values of parameter $W$ in the FC layer.
After the fully connected layer, we add a sigmoid layer to map each response $h_i \in \mathbb{R}$ to $p_i \in [0,1]$, indicating the probability that the pattern appears in the input image:
\begin{equation}
p_i = \frac{1}{1+e^{-h_i}}
\end{equation}

The cost function is defined in Equation \ref{eq:loss}:
\small{
\begin{align}\label{eq:loss}
Loss&= -\frac{1}{N_f} \sum_{i=1}^{N_f} ( \frac{1}{|\mathcal{B}|}  \sum_{j=1}^{|\mathcal{B}|} (1-y_j){\log(1- p_{i,j})} + y_j {\log(p_{i,j})})
\end{align}
}
where $y_j \in \{0,1\}$, $|\mathcal{B}|$ is the size of a mini-batch $\mathcal{B}$ , and $p_{i,j}$ is the response of the $i$-th neuron in the fully connected layer w.r.t. the $j$-th image in the mini batch.
Suppose there are $N_f$ neurons in the fully connected layer.
Then we can get $N_f$ linear combinations of filters from PatternNet by checking the weights of the FC layer.
Each linear combination of filters represents a visual pattern from the given image set.

The intuition of this loss function is to learn multiple visual patterns and use the visual patterns to distinguish positive and negative classes. 
Each neuron in the FC layer selects a few filters to represent a pattern.
If a filter is activated by an image and its corresponding weights in $W$ are nonzero, it will contribute to the output of the FC layer.
The nonzero weights of the parameters of the FC layer control which filters can contribute to the loss function.
The loss function encourages the network to discover filters that are triggered by positive images but not by negative images.
A collection of filters combined together can effectively represent a unique pattern from positive images.
The discovered pattern is representative because most of the positive images are required to trigger the selected filters in the training process.
Furthermore, the discovered pattern is also discriminative because most of the negative images cannot trigger those filters.

After we learn the PatternNet, we can discover the visual patterns from the trained network parameters. For example, the fully connected layer has $N_f*N_c$ dimensional weights, where $N_c$ is the number of filters in the last convolution layer, and $N_f$ is a tunable parameter, which represents the expected number of discovered patterns. We can get $N_f$ combinations of filters by checking the weights of the FC layer in PatternNet. Each combination of filters is able to find a visual pattern from the given image set. To be more specific, we use the following example to explain how to find the visual patterns by PatternNet. We use the similar structure of the convolutional layers from AlexNet to design PatternNet. The last convolutional layer has 256 filters. Each filter produces a $13 \times 13$ feature map. The global max pooling layer has a $13 \times 13$ kernel and produces a single value for each feature map. Therefore, it produces a 256-dimensional vector as the input to the fully connected layer. The fully connected layer has $N_f*N_c$ parameters, in this particular example, we set it to $20 \times 256$, where 256 is determined by the number of filters in the last convolutional layer, and 20 is the expected number of visual patterns that can be discovered from the dataset. For each of the 20 neurons in the FC layer, we check the $N_c$ (which is 256 in AlexNet) parameters, and each parameter correspond to a filter in the last convolutional layer. Since these parameters are sparse, we select the top three parameters, which correspond to three convolutional filters. The select three convolutional filters can be used to detect visual patterns from the image set. We can simply feed an image into the convolutional layers, and check the response map of these selected filters. If all of the filters are activated by the image, it means the image contains a visual pattern defined by these three filters.

We use the deconvolutional neural network architecture (see Fig. \ref{fig:patternnet}) to localize where a visual pattern appears in the input image. The output of the deconvolutional neural network $H_i^{deconv}$ is the derivative of the response value of the $i$-th filter $H_i^{conv}$ w.r.t the pixels in the input images.
\small{
\begin{equation}\label{eq:deconv}
H_{i}^{deconv}(x,y) = \frac{\partial H_i^{conv} }{\partial P}
\end{equation}
}
where $H_i^{deconv}$ is the output of a deconvolutional neural network for the $i^{th}$ filter. It has the same size as the input image $I$. And $P$ is a pixel $(x,y)$ in the input image $I$.
$H_i^{deconv}(x,y)$ is the deconvolutional results at pixel $(x,y)$. It is the partial derivative of $i^{th}$ feature map $w.r.t.$ the pixel $(x,y)$ of the input image $I$.
The value of $H_i^{deconv}(x,y)$ evaluates the impact of input image pixel $(x,y)$ on the $i^{th}$ feature map. The larger value means that this pixel has greater impact. Thus, the deconvolutional results can be used to find which region of the input image contributes to the response map of the given filter. 

A deconvolutional neural network has two passes: a convolution pass and a deconvolution pass. The deconvolution pass shares a similar architecture with the convolution pass, but it performs inverse-like operations, such as pooling vs. unpooling and convolution vs. deconvolution. Unpooling is fairly simple; this operation uses ``switches'' from the pooling layer to recover the feature map prior to the pooling layer. The ``switches'' recover the max value at the original location within each pooling kernel and set all the other locations in the feature map to zero. The non-linear operations (``ReLU'') in the convolution pass are ignored in the deconvolution pass, because the positive values are passed to the next layer without any change and the negative values are set to zero in ReLU. 
The deconvolution operation we use is equivalent to the standard weight update equations during the backpropagation. 
Using the chain-rule, the calculation of partial derivative in Eq. \ref{eq:deconv} is relatively easy. 
In the convolutional neural network, we calculate the output of the $k^{th}$ convolutional layer with kernel $F$ by 
\begin{equation}
H_k = H_{k-1} * F
\end{equation}
For each deconvolutional layer, given the feature map $H^k$ from the $k$-th layer, we use the transpose of the same convolution kernel $F^{T}$ to compute the output of the deconvolutional layer $H^{k-1}$:
\small{
\begin{equation}
\label{eq:deconv_partial}
H^{k-1} = \frac{\partial H^{k}}{\partial F} = H^{k} * F^{T}
\end{equation}
}
The detailed derivation of Equation \ref{eq:deconv_partial} can be found in \cite{zeiler2014visualizing}.
Based on Equation \ref{eq:deconv_partial}, the calculation of Equation \ref{eq:deconv} is equivalent to the backpropagation of a single filter activation.

$H_i^{deconv}$ reflects the impact of each pixel in the input image to the response value of the $i^{th}$ filter. Let $R_i$ be the region of non-zero pixels in $H_i^{deconv}$. Only the pixels in this region contribute to the response value of the filter. The region (patch) of a visual pattern in the input image is:
\small{
\begin{equation}
R_{\mathcal{P}} = \bigcap_{i \in \mathcal{P}} R_i
\end{equation}
}
where $\mathcal{P}$ is a set of filters that define a visual pattern.

\begin{figure}
  \centering
  \includegraphics[width=0.35\textwidth]{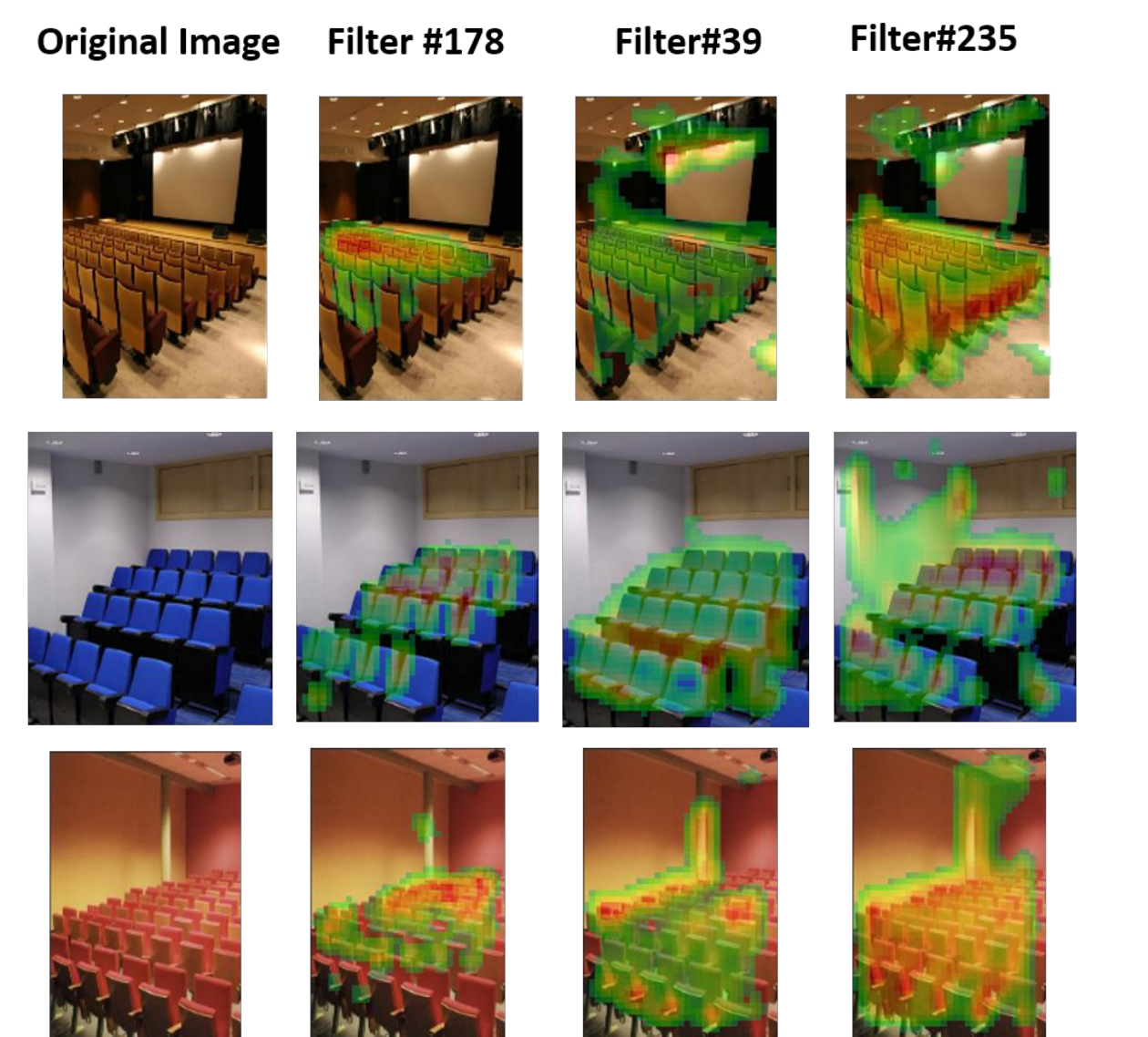}
  \caption{\small{Visualization of the local response magnitude of CNN filters in image patches. The heatmap shows different filters are selective to different \textit{latent} patterns.}}
  \label{fig:heatmap}
\end{figure}

To illustrate how a set of filters finds a visual pattern, we generate heat maps to visualize the local response region of CNN filters in Fig. \ref{fig:heatmap}.
Fig. \ref{fig:heatmap} shows a ``seat'' pattern is found in the image collection with filter \#178, \#39 and \#235. For each image, all the filters are activated at the same location. The figure clearly shows which region activates the filters and it is obvious that ``seat'' is the target of this pattern.

\begin{figure*}
  \centering
  \includegraphics[width=0.9\textwidth]{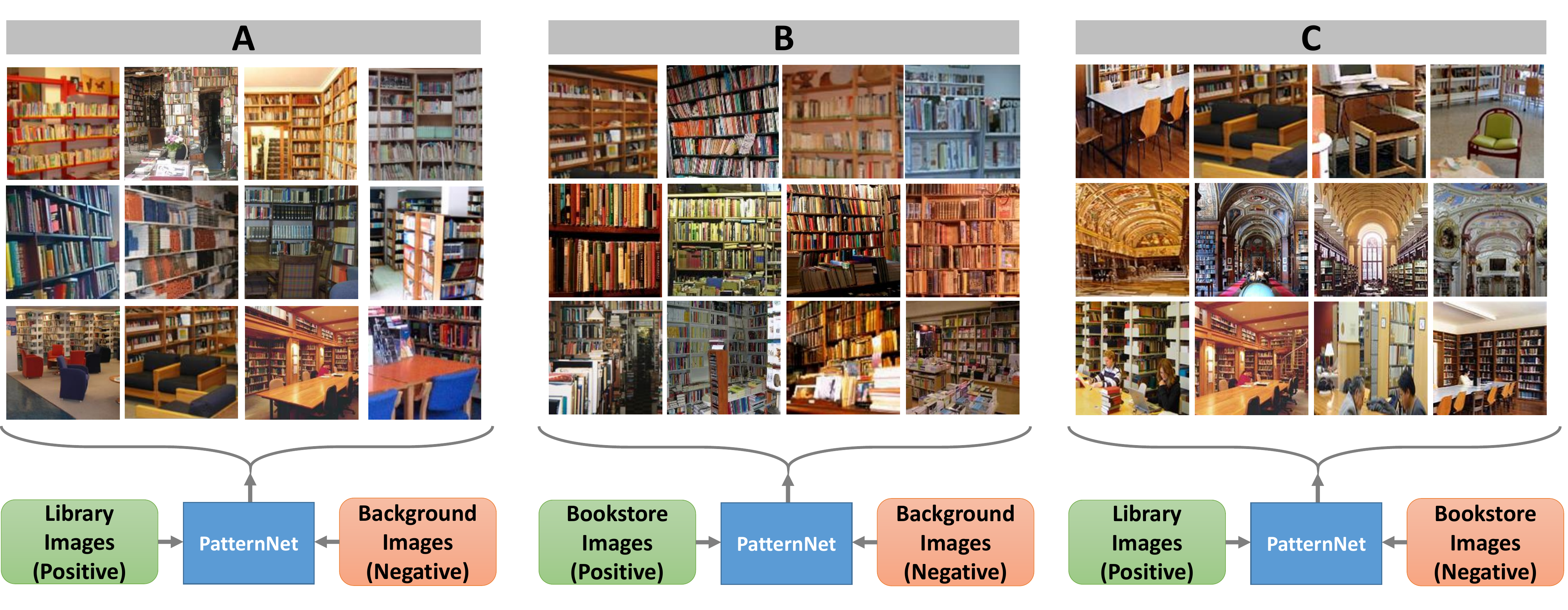}
  \caption{Find discriminative patterns by using different negative images.}
  \label{fig:library}
\end{figure*}

\subsection{Finding Discriminative Patterns from a Category w.r.t Different Reference Images}
In contrast to \textit{Representative}, \textit{Discriminative} is a relative concept.
For example, the visual pattern ``book shelf'' is a \textit{discriminative} pattern in the ``library'' images w.r.t. to the random background images\footnote{e.g. random images downloaded from Flickr, or random images selected from all the other categories}.
But it is not a \textit{discriminative} pattern if we use images from the ``bookstore'' category as reference images since there are also many ``bookshelf'' visual instances in the ``bookstore'' images.

As shown in Fig. \ref{fig:library} (A) and (B), we use random background images as reference images and find visual patterns from ``library'' images and ``bookstore'' images.
We find that both ``books'' and ``book shelf'' visual patterns are \textit{discriminative} patterns for the two categories w.r.t. random background images.
But if we want to find the \textit{discriminative} patterns from ``library'' images w.r.t. ``bookstore'' images, as shown in Figure \ref{fig:library} (C), the unique patterns like  ``chairs'', ``the hall'', and ``reading desks'' are discovered, instead of patterns like ``bookshelf'', which are shared between the two categories.

This characteristic demonstrates that PatternNet has the capability to find the ``true'' discriminative patterns from a set of images w.r.t different reference images. This is quite a useful feature. For example, given a set of images taken in London and another set of images taken in New York, we can use this algorithm to find out what are the unique visual characteristics in London.

\section{Experiment}
\subsection{Subjective Evaluation}
\begin{figure*}
  \centering
  \includegraphics[width=0.75\textwidth]{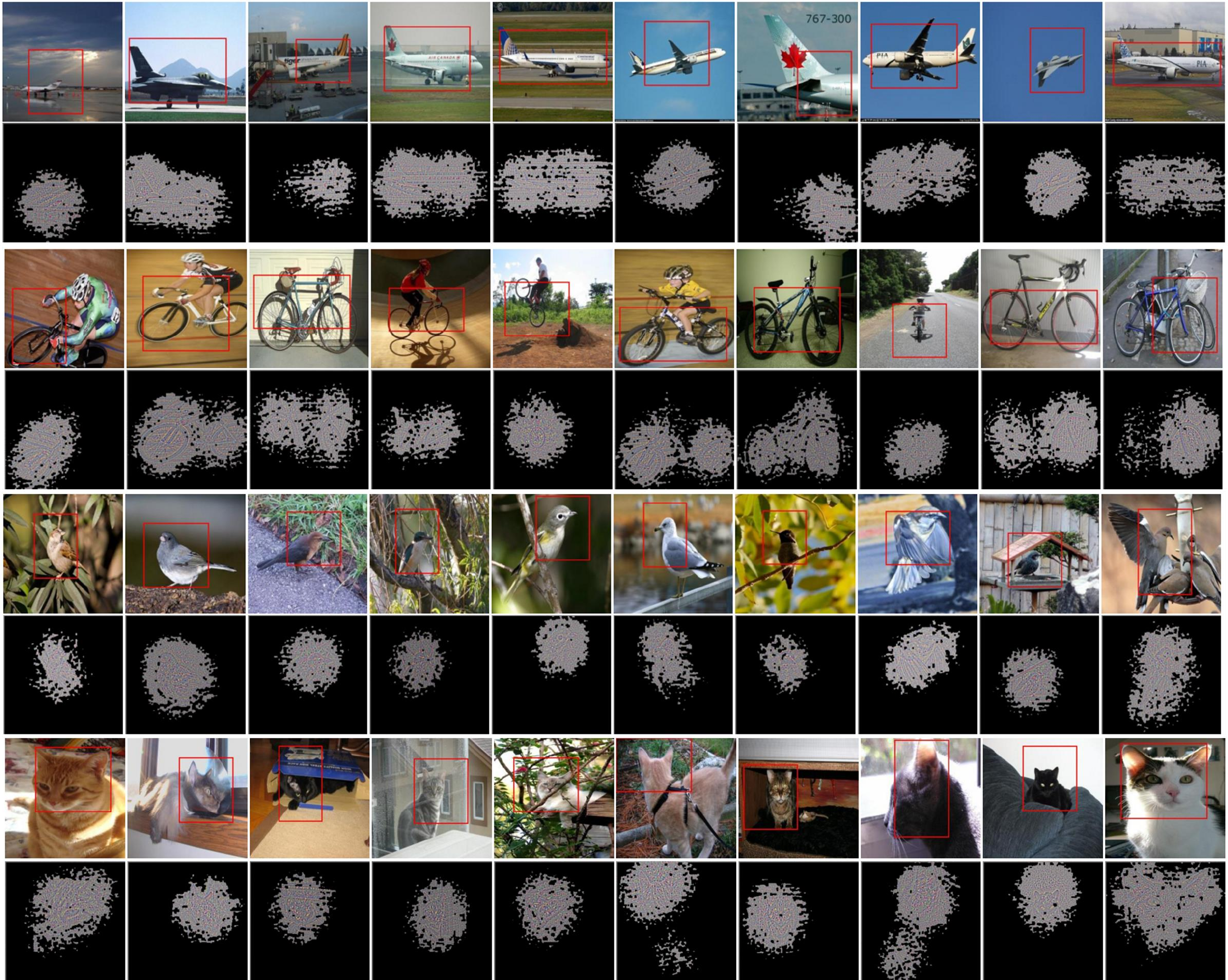}
  \caption{Randomly selected visual patterns discovered by PatternNet from a variety of datasets. The mask images show the localization results of each visual pattern.}
  \label{fig:examplepattern}
\end{figure*}

To demonstrate the performance of PatternNet, we first present in Fig. \ref{fig:examplepattern} some randomly selected visual patterns discovered from a variety of datasets, including VOC2007, MSCOCO and CUB-Bird-200. The mask images are generated from a deconvolutional neural network \cite{zeiler2014visualizing}, which demonstrates the ability of PatternNet to discover and visualize visual patterns in images. 
We can clearly see that PatternNet is able to find and localize the intricate visual patterns from different datasets.

\subsection{Objective Evaluation}
It is not easy to directly evaluate visual pattern mining works due to the lack of well-annotated datasets for this task.
Some previous works use the image classification task as a proxy to evaluate visual pattern mining results \cite{lilsh15cvpr}. In this paper, we also conduct experiments for image classification as a proxy to evaluate our work.
We compare the PatternNet model with several baseline approaches across a wide variety of datasets.
We also compare with several state-of-the-art visual pattern mining or visual instance mining works for scene classification and fine-grained image classification. It is important to note that we use the classification tasks as a proxy to evaluate the quality of visual pattern mining methods. We are not aiming to outperform all the state-of-the-art image classification methods. Instead, we believe that our visual pattern mining approach could be used to improve current image classification methods. Thus, in the following experiments, we only compare our proposed PatternNet with other state-of-the-art visual pattern mining methods on image classification tasks.

We believe that it is not sufficient to evaluate the quality of visual patterns by solely using image classification. Ideally, we should use a dataset with all possible patterns labeled by human beings to measure the precision, recall, and F-score of the discovered visual patterns. However, it is almost impossible to label a large-scale dataset due to the difficulty and amount of human labor required. Instead, we notice that the currently available datasets for object detection can be used to evaluate pattern mining works. This is because the labeled objects for the detection task in those datasets are indeed ``discriminative'' and ``representative'' visual patterns. If the pattern mining algorithm is robust, those objects should be detected and found by our approach. In this paper, we follow the evaluation metric proposed in \cite{uijlings2013selective} to directly evaluate the quality of visual patterns discovered by our work. \cite{uijlings2013selective} developed a technique to propose a bounding box for candidate objects from images. In their paper, they use Mean Average Best Overlap (MABO) as the evaluation metric. $ABO$ for a specific class $c$ is defined as follows:
\begin{equation}
ABO = \frac{1}{|G^c|} \sum_{g_i^{c} \in G^c} \max_{l_j \in L} Overlap(g_i^c, l_j)
\end{equation}
\begin{equation}
Overlap(g_i^c, l_j) = \frac{area(g_i^c) \cap area(l_j)} {area(g_i^c) \cup area(l_j)}
\end{equation}
where $g_i^c \in G^c$ is the ground truth annotation and $L$ is the bounding box of the detected visual patterns.

\subsection{Using Image Classification as a Proxy to Evaluate PatternNet}
We follow similar settings to those used by most existing pattern mining or instance mining works to evaluate our approach to image classification. That is, after we discover the visual patterns from the training images, we use the visual patterns to extract a middle-level feature representation for both training and test images. Then, the middle-level feature representations are used in the classification task.
As our visual patterns are discovered and represented by a set of filters in the last convolution layer, it is easy and natural for us to integrate the pattern detection, training, and testing phrases together in one neural network.
To do this, we add a static fully connected layer on top of the last convolution layer.
The parameters of this FC layer are manually initialized using the visual patterns learned in PatternNet.
During the training phase, we freeze this static FC layer by setting the learning rate to zero.
Assume that we have $N_f$ unique visual patterns for a dataset. Then the FC layer has $N_f$ dimensional outputs. Each dimension collects the response from the filters associated with one visual pattern.
After scaling and normalization, the value of each dimension represents the detection score of a visual pattern from an input image.
On top of this FC layer, a standard FC layer and softmax layer are used to obtain classification results.

\subsubsection{Baseline Comparison}
We compare PatternNet with several baseline approaches, including: 1) use the response of fc7 layer of a CNN \cite{Simonyan14c} as image feature representation and train a multi-class SVM model for image classification, 2) use the response of the pool5 layer from a CNN \cite{Simonyan14c} as image feature representation and train a multi-class SVM model for image classification.
We simply use a fully connected layer and a softmax layer on top of the PatternNet architecture to classify images.
The results can be found in Table \ref{tab:baseline}. PatternNet outperforms the baseline approaches that directly use the response from the pool5 and fc7 layers as image features. The response of each neuron in the last fully connected layer in PatternNet indicates whether the input image has a certain visual pattern. The results in Table \ref{tab:baseline} prove that the selected visual patterns are discriminative and can be used as effective image features for classification.
\small{
\begin{table}
\centering
\caption{Comparison with the baseline approaches on some popular benchmark datasets. PatternNet uses the same structure of convolutional layers as in VGG19, and uses their pre-trained model to initialize the convolutional layers}
\begin{tabular}{c|ccccccc}
\hline
\hline
Method  &  MITIndoor	& CUB-BIRD-200 & StanfordDogs \\ \hline
VGG19 fc7 + SVM   & 67.6   &   54.6    &  78.1        \\
VGG19 pool5 + SVM   & 61.6             &  40.7   &67.1 \\
PatternNet   & 75.3        &   70.0   & 83.1    \\
\hline
\end{tabular}
\label{tab:baseline}
\end{table}
}

\small{
\begin{table}
\centering
\caption{Scene classification results for MITIndoor dataset. PatternNet uses the pre-trained AlexNet to initialize the convolutional layers to achieve a fair comparison.}
\begin{tabular}{lc}
\hline
\hline
Method                       & Accuracy (\%)           \\ \hline
ObjectBank \cite{li2010object}                                  & 37.60                  \\
Discriminative Patch \cite{singh2012unsupervised}                     & 38.10             \\
BoP \cite{juneja2013blocks}                                & 46.10                  \\
HMVC \cite{li2013harvesting}                                  & 52.30                 \\
Discriminative Part \cite{sun2013learning}                 & 51.40                  \\
MVED \cite{doersch2013mid}                  & 66.87                  \\
MDPM \cite{lilsh15cvpr}                                     & 69.69                  \\
PatternNet                                      &  71.30       \\ \hline
\\
\end{tabular}
\label{tab:MITIndoor}
\end{table}
}

\subsubsection{Scene Classification}
We use the MITIndoor dataset for scene classification, which has 67 classes of indoor images.
We follow the split of training and test images as in \cite{quattoni2009recognizing}: approximately 80 training images and 20 test images per category.
About 20 visual patterns are discovered by PatternNet on training images for each category.
For each indoor scene category, we use its 80 images as positive samples and the images from other categories as negative samples to train the PatternNet model to discover visual patterns.
The convolutional layers are initialized by a pre-trained CNN model \cite{krizhevsky2012imagenet}, and frozen during the training phase.
PatternNet converges within approximately 100 iterations, which takes about 1-2 mins on a workstation with GTX Titan X GPU.
After this procedure, we find approximately 20-30 unique patterns per scene category.
Note that the number of discovered patterns is controllable by using different dimensions of parameters in the fully connected layer of PatternNet.
From Table \ref{tab:MITIndoor}, we can see that PatternNet outperforms the state-of-the-art works. Compared with MDPM, we directly modify the current CNN architecture to perform the scene classification task,
while their approach has to sample image patches from test images and produce the middle-level feature representation for classification.
In addition to enhanced performance, this allows our approach to provide an order of magnitude speedup compared with MDPM.

\begin{table}
\centering
\caption{Fine-grained object classification results on Stanford dogs dataset}
\begin{tabular}{lc}
\hline
\hline
 \ \ \ \ \ \ Method    \ \  \ \ \ \ \ \ \ \ \  \ \ \ \ \ \ \ \ \             & Accuracy (\%)           \\ \hline
Alignments \cite{gavves2014local}                           & 36.8                  \\
GMTL \cite{pu2014looks}                                  & 39.3                  \\
Symb \cite{chai2013symbiotic}                           & 45.6                 \\
Unsupervised Alignments \cite{gavves2013fine}  & 50.1 \\
SPV \cite{chen2015selective}        & 52.0 \\
Google LeNet ft \cite{szegedy2015going}  & 75.0 \\
Attention \cite{sermanet2014attention} & 76.8 \\
\textbf{PatternNet}                              &  \textbf{83.1}        \\ \hline \\
\end{tabular}
\label{tab:dogs}
\end{table}

\subsubsection{Fine-grained Object Classification}
Recently, fine-grained image classification has attracted much attention in visual recognition.
Compared with traditional image classification problems (dogs vs. cars, buildings vs. people), the fine-grained image classification (Labrador vs. Golden Retriever, Husky vs. Samoyed) is a much more challenging problem, since it requires attention to detailed visual features in order to distinguish the fine-grained categories.
Similar to the scene classification task, our insight is that discriminative patterns are able to capture the information from local parts of the image/object.
With PatternNet, the discriminative patterns of each fine-grained class can be effectively discovered.
As is the property of discriminative patterns, the patterns from one category rarely appear in other categories.
Hence, such patterns have a great potential to improve the fine-grained object classification task.

We evaluate our approach on two popular datasets for the fine-grained image classification task: CUB-bird 200 and Stanford Dogs.
The CUB-200 dataset has 200 classes of different birds, and the Stanford Dogs dataset has 120 categories of different dogs.
We follow the suggested training-test split from the original dataset and compare our results with some state-of-the-art works as listed in Tables \ref{tab:dogs} and \ref{tab:bird}.
The parameters of the convolution layers are imported from a pre-trained CNN model \cite{Simonyan14c} without any fine-tuning.
During any of our training phases, we do not fine-tune any parameters from convolution layers in order to prevent over-fitting.
The only parameters we learned for PatternNet are the fully connected layer as the indicator of linear combination of convolution filters.
The training phase for discovering patterns is stopped after a few hundreds of iterations when the training loss is stable.
We notice that some recent works reported significantly high performance by leveraging the manually labeled bounding box information, such as 82.0\% reported by PD  \cite{krause2015fine} on the CUB-200 dataset.
In our approach, we use neither manually labeled bounding boxes nor parts information.
The purpose of this experiment is to evaluate the quality of the discovered visual patterns.
It is important to evaluate our approach on the whole image instead of clean objects given by the manually labeled bounding box.
Thus, we only compare with the approaches that do not use the manually labeled bounding box information.
We also do not compare with works that use additional images to fine-tune a CNN model, since those works involve additional training data and thus are not a fair comparison.
From tables \ref{tab:dogs} and \ref{tab:bird}, we can see the clear advantage of our approach compared with the state-of-the-art works on the same experiment setup.

\subsection{Evaluate PatternNet on Object Proposal Task}
As we have discussed before, visual pattern mining technology can be used for the object proposal task. The main difference between pattern mining and traditional object proposal methods (\cite{uijlings2013selective} \cite{alexe2012measuring}) is that we do not need manually labeled object bounding boxes to train the model. Our algorithm directly mines the regularity from the given image set and finds important (``discriminative'' and ``representative'') objects in images. Also, instead of proposing thousands of object bounding box proposals in \cite{uijlings2013selective}, we only generate tens of object bounding boxes with a much higher accuracy. As \cite{uijlings2013selective} is widely used in many research works for pre-processing images, we compare our approach with the state-of-the-art object proposal works and show the results in Table \ref{tab:recall}. Our advantage is that we propose far fewer bounding boxes than the traditional object proposal works. We compare the recall rate and MABO reported in the literature when approximately 100 bounding boxes are proposed by those works.
The results show that the PatternNet outperforms the other works with a substantially lower number of proposed bounding boxes.

\begin{table}
\centering
\caption{Fine-grained classification results on CUB-200 dataset}
\begin{tabular}{lc}
\hline
\hline
\ \ \ \ \ \ \ \ Method  \ \ \ \ \ \ \ \ \ \ \ \ \ \ \ \ & Accuracy (\%)           \\ \hline
GMTL \cite{pu2014looks}                           & 44.2                 \\
SPV \cite{chen2015selective}        & 48.9  \\
Alignments \cite{gavves2014local}                                 & 53.6                  \\
POOF \cite{berg2013poof}                                       & 56.8                  \\
R-CNN \cite{girshick2014rich}                          & 58.8                  \\
Symb \cite{chai2013symbiotic}                                    & 59.4                  \\
PB-R-CNN \cite{zhang2014part}       &  65.9  \\
\textbf{PatternNet}                             &  \textbf{70.0}        \\ \hline  \\
\end{tabular}

\label{tab:bird}
\end{table}

\begin{table}
\centering
\caption{Comparison of recall rate and MABO for a variety of methods on the Pascal VOC 2007 test set. For PatternNet, we propose about 5 bounding boxes per image. For the other methods, we compare their reported number when about 100 bounding boxes are proposed.}
\begin{tabular}{lccc}
\hline
\hline
\ \ \ \ \ \ \ \ \  Method  &   \ Recall & MABO & \small{ \# proposed BB}  \\ \hline
Sliding window search \cite{harzallah2009combining}            & 0.75       & -    &  100       \\
Jumping windows \cite{vedaldi2009multiple} &  0.60 & - & 100\\
Objectness \cite{alexe2012measuring} & 0.77 & 0.66 & 100 \\
The boxes around the regions \cite{carreira2010constrained} & 0.70  & 0.63 & 100 \\
The boxes around the regions \cite{endres2010category} & 0.74 & 0.65 & 100 \\
Selective Search \cite{uijlings2013selective} & 0.74 &  0.63 & 100\\
\textbf{PatternNet}                              &  \textbf{0.86}  &  \textbf{0.77} & 5   \\ \hline   \\
\end{tabular}
\label{tab:recall}
\end{table}

\section{Conclusion}
In this paper, we have presented a novel neural network architecture called PatternNet for discovering visual patterns from image collections.
PatternNet leverages the capability of the convolutional layers in a CNN, where each filter normally has consistent response to certain high-level visual patterns.
This property is used to discover discriminative and representative visual patterns by using a specially designed fully connected layer and a lost function to
find a sparse combinations of filters, which have strong responses to the patterns in images from the target category and weak responses to images from the rest of the categories.
We conducted experimental evaluation on both the scene classification task and the fine-grained object classification task. The evaluation result shows that the discovered visual patterns by PatternNet are both representative and discriminative.
We believe that PatternNet has shown promising performance in automatically discovering the useful portions of an image and enables advanced computer vision applications without expensive bounding box based labeling of datasets.

%% file: main.bbl
\begin{thebibliography}{10}

\bibitem{agrawal1993mining}
R.~Agrawal, T.~Imieli{\'n}ski, and A.~Swami.
\newblock Mining association rules between sets of items in large databases.
\newblock In {\em ACM SIGMOD Record}, volume~22, pages 207--216. ACM, 1993.

\bibitem{alexe2012measuring}
B.~Alexe, T.~Deselaers, and V.~Ferrari.
\newblock Measuring the objectness of image windows.
\newblock {\em Pattern Analysis and Machine Intelligence, IEEE Transactions
  on}, 34(11):2189--2202, 2012.

\bibitem{berg2013poof}
T.~Berg and P.~N. Belhumeur.
\newblock Poof: Part-based one-vs.-one features for fine-grained
  categorization, face verification, and attribute estimation.
\newblock In {\em Computer Vision and Pattern Recognition (CVPR), 2013 IEEE
  Conference on}, pages 955--962. IEEE, 2013.

\bibitem{carreira2010constrained}
J.~Carreira and C.~Sminchisescu.
\newblock Constrained parametric min-cuts for automatic object segmentation.
\newblock In {\em Computer Vision and Pattern Recognition (CVPR), 2010 IEEE
  Conference on}, pages 3241--3248. IEEE, 2010.

\bibitem{chai2013symbiotic}
Y.~Chai, V.~Lempitsky, and A.~Zisserman.
\newblock Symbiotic segmentation and part localization for fine-grained
  categorization.
\newblock In {\em Computer Vision (ICCV), 2013 IEEE International Conference
  on}, pages 321--328. IEEE, 2013.

\bibitem{chen2015selective}
G.~Chen, J.~Yang, H.~Jin, E.~Shechtman, J.~Brandt, and T.~X. Han.
\newblock Selective pooling vector for fine-grained recognition.
\newblock In {\em Applications of Computer Vision (WACV), 2015 IEEE Winter
  Conference on}, pages 860--867. IEEE, 2015.

\bibitem{doersch2013mid}
C.~Doersch, A.~Gupta, and A.~A. Efros.
\newblock Mid-level visual element discovery as discriminative mode seeking.
\newblock In {\em Advances in Neural Information Processing Systems}, pages
  494--502, 2013.

\bibitem{endres2010category}
I.~Endres and D.~Hoiem.
\newblock Category independent object proposals.
\newblock In {\em Computer Vision--ECCV 2010}, pages 575--588. Springer, 2010.

\bibitem{gavves2013fine}
E.~Gavves, B.~Fernando, C.~G. Snoek, A.~W. Smeulders, and T.~Tuytelaars.
\newblock Fine-grained categorization by alignments.
\newblock In {\em Proceedings of the IEEE International Conference on Computer
  Vision}, pages 1713--1720, 2013.

\bibitem{gavves2014local}
E.~Gavves, B.~Fernando, C.~G. Snoek, A.~W. Smeulders, and T.~Tuytelaars.
\newblock Local alignments for fine-grained categorization.
\newblock {\em International Journal of Computer Vision}, 111(2):191--212,
  2014.

\bibitem{girshick2014rich}
R.~Girshick, J.~Donahue, T.~Darrell, and J.~Malik.
\newblock Rich feature hierarchies for accurate object detection and semantic
  segmentation.
\newblock In {\em Computer Vision and Pattern Recognition (CVPR), 2014 IEEE
  Conference on}, pages 580--587. IEEE, 2014.

\bibitem{hariharan2015hypercolumns}
B.~Hariharan, P.~Arbel{\'a}ez, R.~Girshick, and J.~Malik.
\newblock Hypercolumns for object segmentation and fine-grained localization.
\newblock In {\em Proceedings of the IEEE Conference on Computer Vision and
  Pattern Recognition}, pages 447--456, 2015.

\bibitem{harzallah2009combining}
H.~Harzallah, F.~Jurie, and C.~Schmid.
\newblock Combining efficient object localization and image classification.
\newblock In {\em Computer Vision, 2009 IEEE 12th International Conference on},
  pages 237--244. IEEE, 2009.

\bibitem{juneja2013blocks}
M.~Juneja, A.~Vedaldi, C.~Jawahar, and A.~Zisserman.
\newblock Blocks that shout: Distinctive parts for scene classification.
\newblock In {\em Computer Vision and Pattern Recognition (CVPR), 2013 IEEE
  Conference on}, pages 923--930. IEEE, 2013.

\bibitem{krause2015fine}
J.~Krause, H.~Jin, J.~Yang, and L.~Fei-Fei.
\newblock Fine-grained recognition without part annotations.
\newblock In {\em Proceedings of the IEEE Conference on Computer Vision and
  Pattern Recognition}, pages 5546--5555, 2015.

\bibitem{krizhevsky2012imagenet}
A.~Krizhevsky, I.~Sutskever, and G.~E. Hinton.
\newblock Imagenet classification with deep convolutional neural networks.
\newblock In {\em Advances in neural information processing systems}, pages
  1097--1105, 2012.

\bibitem{li2010object}
L.-J. Li, H.~Su, L.~Fei-Fei, and E.~P. Xing.
\newblock Object bank: A high-level image representation for scene
  classification \& semantic feature sparsification.
\newblock In {\em Advances in neural information processing systems}, pages
  1378--1386, 2010.

\bibitem{li2013harvesting}
Q.~Li, J.~Wu, and Z.~Tu.
\newblock Harvesting mid-level visual concepts from large-scale internet
  images.
\newblock In {\em Computer Vision and Pattern Recognition (CVPR), 2013 IEEE
  Conference on}, pages 851--858. IEEE, 2013.

\bibitem{lilsh15cvpr}
Y.~Li, L.~Liu, C.~Shen, and A.~van~den Hengel.
\newblock Mid-level deep pattern mining.
\newblock In {\em CVPR}, 2015.

\bibitem{lowe1999object}
D.~G. Lowe.
\newblock Object recognition from local scale-invariant features.
\newblock In {\em Computer vision, 1999. The proceedings of the seventh IEEE
  international conference on}, volume~2, pages 1150--1157. Ieee, 1999.

\bibitem{parizi2014automatic}
S.~N. Parizi, A.~Vedaldi, A.~Zisserman, and P.~Felzenszwalb.
\newblock Automatic discovery and optimization of parts for image
  classification.
\newblock {\em arXiv preprint arXiv:1412.6598}, 2014.

\bibitem{pu2014looks}
J.~Pu, Y.-G. Jiang, J.~Wang, and X.~Xue.
\newblock Which looks like which: Exploring inter-class relationships in
  fine-grained visual categorization.
\newblock In {\em Computer Vision--ECCV 2014}, pages 425--440. Springer, 2014.

\bibitem{quattoni2009recognizing}
A.~Quattoni and A.~Torralba.
\newblock Recognizing indoor scenes.
\newblock In {\em Computer Vision and Pattern Recognition, 2009. CVPR 2009.
  IEEE Conference on}, pages 413--420. IEEE, 2009.

\bibitem{sermanet2014attention}
P.~Sermanet, A.~Frome, and E.~Real.
\newblock Attention for fine-grained categorization.
\newblock {\em arXiv preprint arXiv:1412.7054}, 2014.

\bibitem{Simonyan14c}
K.~Simonyan and A.~Zisserman.
\newblock Very deep convolutional networks for large-scale image recognition.
\newblock {\em arXiv preprint arXiv:1409.1556}, 2014.

\bibitem{singh2012unsupervised}
S.~Singh, A.~Gupta, and A.~Efros.
\newblock Unsupervised discovery of mid-level discriminative patches.
\newblock {\em Computer Vision--ECCV 2012}, pages 73--86, 2012.

\bibitem{sun2013learning}
J.~Sun and J.~Ponce.
\newblock Learning discriminative part detectors for image classification and
  cosegmentation.
\newblock In {\em Computer Vision (ICCV), 2013 IEEE International Conference
  on}, pages 3400--3407. IEEE, 2013.

\bibitem{szegedy2015going}
C.~Szegedy, W.~Liu, Y.~Jia, P.~Sermanet, S.~Reed, D.~Anguelov, D.~Erhan,
  V.~Vanhoucke, and A.~Rabinovich.
\newblock Going deeper with convolutions.
\newblock In {\em Proceedings of the IEEE Conference on Computer Vision and
  Pattern Recognition}, pages 1--9, 2015.

\bibitem{uijlings2013selective}
J.~R. Uijlings, K.~E. van~de Sande, T.~Gevers, and A.~W. Smeulders.
\newblock Selective search for object recognition.
\newblock {\em International journal of computer vision}, 104(2):154--171,
  2013.

\bibitem{vedaldi2009multiple}
A.~Vedaldi, V.~Gulshan, M.~Varma, and A.~Zisserman.
\newblock Multiple kernels for object detection.
\newblock In {\em Computer Vision, 2009 IEEE 12th International Conference on},
  pages 606--613. IEEE, 2009.

\bibitem{zeiler2014visualizing}
M.~D. Zeiler and R.~Fergus.
\newblock Visualizing and understanding convolutional networks.
\newblock In {\em Computer Vision--ECCV 2014}, pages 818--833. Springer, 2014.

\bibitem{zhang2014part}
N.~Zhang, J.~Donahue, R.~Girshick, and T.~Darrell.
\newblock Part-based r-cnns for fine-grained category detection.
\newblock In {\em Computer Vision--ECCV 2014}, pages 834--849. Springer, 2014.

\bibitem{zhang2014scalable}
W.~Zhang, H.~Li, C.-W. Ngo, and S.-F. Chang.
\newblock Scalable visual instance mining with threads of features.
\newblock In {\em Proceedings of the ACM International Conference on
  Multimedia}, pages 297--306. ACM, 2014.

\end{thebibliography}
